\documentclass{article}

\usepackage{threeparttable,booktabs,array,booktabs,multirow,fontsize,arydshln,xcolor,setspace,chngcntr,subcaption,placeins}
\usepackage{spconf,amsmath,graphicx,hyperref}

\newcommand{\lightit}[1]{{\fontseries{l}\fontshape{it}\selectfont #1}}

\title{SCHK-HTC: Sibling Contrastive Learning with Hierarchical Knowledge-aware Prompt Tuning For Hierarchy Text Classification}
\name{
Ke Xiong\textsuperscript{1}\quad
Qian Wu\textsuperscript{2}\quad
Wangjie Gan\textsuperscript{1}\quad
Yuke Li\textsuperscript{2}\quad
Xuhong Zhang\textsuperscript{1,*}\quad
\thanks{*Corresponding author: Xuhong Zhang (zhangxuhong@zju.edu.cn)}}

\address{
$^{1}$School of Software Technology, Zhejiang University, China\\
$^{2}$NetEase Yidun AI Lab, Hangzhou, China
}
\begin{document}
    \maketitle
    \ninept
    \begin{abstract}
        Few-shot Hierarchical Text Classification (few-shot HTC) is a challenging task that involves mapping texts to a predefined tree-structured label hierarchy under data-scarce conditions. While current approaches utilize structural constraints from the label hierarchy to maintain parent-child prediction consistency, they face a critical bottleneck, the difficulty in distinguishing semantically similar sibling classes due to insufficient domain knowledge. We introduce an innovative method named \textbf{S}ibling \textbf{C}ontrastive Learning with \textbf{H}ierarchical \textbf{K}nowledge-aware Prompt Tuning for few-shot \textbf{HTC} tasks (\textbf{SCHK-HTC}). Our work enhances the model's perception of subtle differences between sibling classes at deeper levels, rather than just enforcing hierarchical rules. Specifically, we propose a novel framework featuring two core components: a hierarchical knowledge extraction module and a sibling contrastive learning mechanism. This design guides  model to encode discriminative features at each hierarchy level, thus improving the separability of confusable classes. Our approach achieves superior performance across three benchmark datasets, surpassing existing state-of-the-art methods in most cases. Our code is available at \url{https://github.com/happywinder/SCHK-HTC}.
    \end{abstract}
    \begin{keywords}
        hierarchical text classification, prompt tuning, contrastive learning, knowledge graph 
    \end{keywords}
    \vspace{-0.5em}
    \section{Introduction}
    \vspace{-0.5em} 
    \label{sec:intro}
    Hierarchical Text Classification (HTC), a specialized form of multi-label text classification, has found wide-ranging applications \cite{mao2019hierarchical} in numerous real-world scenarios, such as news topic categorization \cite{lewis2004rcv1} and academic paper classification \cite{kowsari2017hdltex}. Few-shot HTC extends this task, presenting even greater challenges. The core objective of few-shot HTC is to accurately classify texts or documents from the coarsest to the finest granularity within a class hierarchy, given an extremely limited number of samples \cite{ji2023hierarchical,chen2024retrieval,cui-etal-2022-prototypical}. 

    With the advent and proliferation of Pre-trained Language Models (PLMs) \cite{devlin2019bert} , the prompt-tuning paradigm \cite{lester2021power} , which employs PLMs as text encoders, has emerged as a dominant research trend \cite{wang2022hpt} . This approach effectively bridges the gap between the pretraining objectives of PLMs and the requirements of downstream tasks. Early prominent HTC methods \cite{chen2021hierarchy,zhou2020hierarchy} utilized graph neural networks \cite{sinha2018hierarchical} to encode the label taxonomy . While effective, these approaches are inherently data-intensive and perform poorly in few-shot scenarios. HierVerb \cite{ji2023hierarchical} introduced a paradigm shift by replacing explicit label hierarchy encoder with a contrastive learning \cite{chen2020simple} objective. This approach proved highly effective, setting new SOTA performance on serveral datasets.  Nevertheless, pulling lower levels' label embedding increases their representational overlap and thus exacerbates confusion, ultimately hindering performance.  This highlights a critical limitation of such approaches: \textit{as classification descends to deeper levels of the hierarchy, the semantic differences between labels become increasingly subtle}, making them difficult to distinguish based solely on the text. This amplifies the need for external knowledge. K-HTC \cite{liu2023enhancing} incorporates Knowledge Graph (KG) \cite{speer2017conceptnet} to provide domain knowledge, aiming to mitigate the interference from general-purpose pre-training data. However, its knowledge utilization is not hierarchical and lack of a mechanism to effectively fuse label semantics with domain-specific knowledge. Furthermore, its performance in low-resource settings was not analyzed. DCL \cite{chen2024retrieval} leverages an external knowledge base through retrieval-augmented generation \cite{lewis2020retrieval} and large language model (LLM) \cite{brown2020language} , achieving impressive performance gains. However, this approach suffers from two significant drawbacks: a massive parameter count that increases computational costs, and a heavy reliance on extensive annotated data for in-context learning \cite{xie2021explanation}. Thus, the challenge of achieving effective discrimination between sibling labels at deeper levels, especially under low-resource constraints, constitutes a central and unresolved issue.
    \begin{figure}[t]
        \centering
        \captionsetup[subfloat]{justification=centering}
        \hspace{-13pt} 
        \subfloat[Classification Acc(\%) on\\ WOS (Layer-2) ]{
            \includegraphics[width=0.5\linewidth]{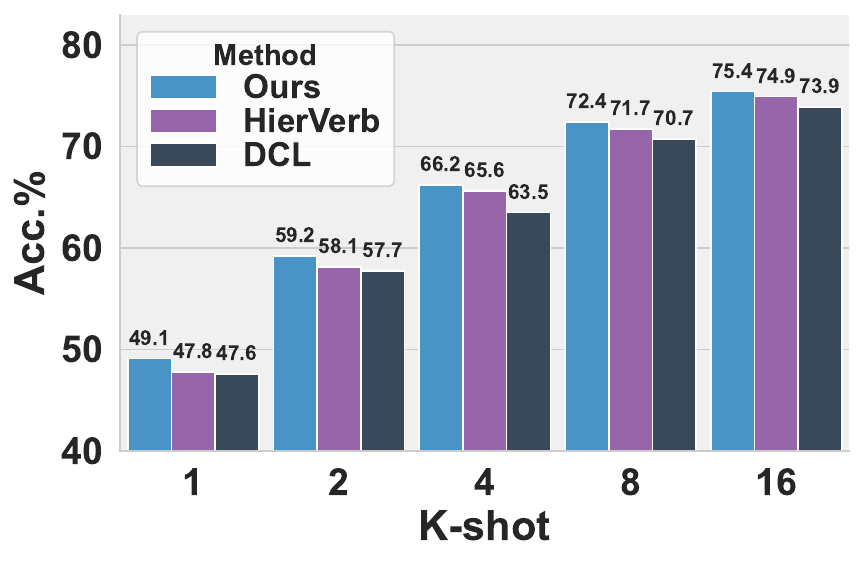}
            \label{fig:wos_2_level}
        }
        \hspace{-10pt} 
        \subfloat[Classification Acc(\%) on\\ DBpedia (Layer-3) ]{
            \includegraphics[width=0.5\linewidth]{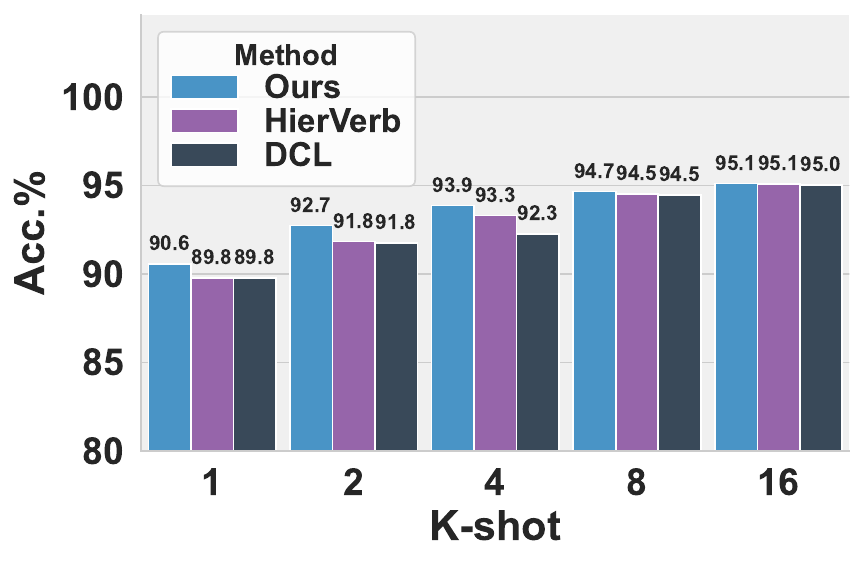}
            \label{fig:dbp_3_level}
        }
        \vspace{-1pt}
        \caption{Classification Acc.(\%) on deepest level of WOS and DBpedia dataset.}
        \vspace{-5pt}
        \label{fig:Deeper Level}
        \vspace{-13pt}
    \end{figure}
    \begin{figure*}[t]
        \centering
        \includegraphics[width=0.95\textwidth]{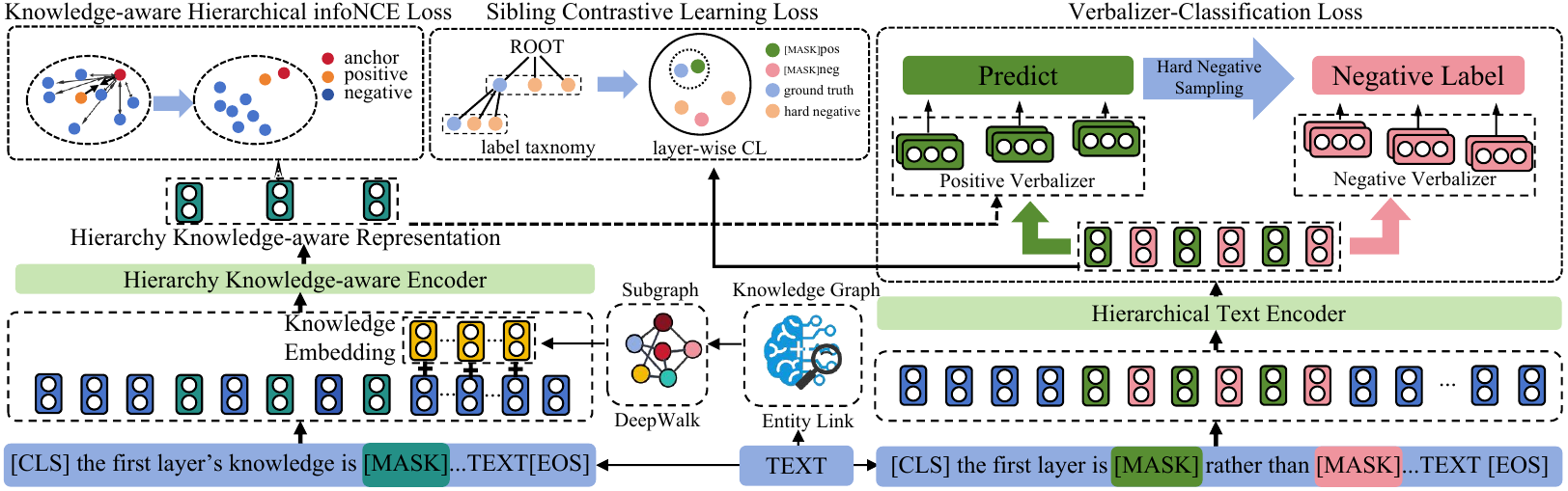} 
        \vspace{-10pt}
        \caption{The overall architecture of the proposed sibling contrastive learning with hierarchical knowledge-aware prompt-tuning (SCHK-HTC) framework.}
        \label{fig:1}
        \vspace{-10pt}
    \end{figure*}
    Motivated by these observations, we propose a novel framework to tackle these challenges through \textbf{two core innovations}. First, to compensate for the scarcity of domain knowledge, we introduce a mechanism to extract hierarchical knowledge features from KG. This provides the model with structured, level-aware context crucial for classification in data-limited settings. Second, to address the ambiguity among fine-grained classes, we employ contrastive learning objective specifically on sibling labels. This forces the model to learn subtle yet critical distinctions between semantically similar categories. Together, these two components enable our model to learn more discriminative representations for effective few-shot HTC.
    The main contributions of this paper are summarized as follows: (1) We propose a novel hierarchical knowledge-aware contrastive learning method based on prompt tuning. (2) We integrate KG into the few-shot HTC to alleviate the issue of insufficient domain knowledge, and employ contrastive learning to further address the problem of high semantic similarity among sibling classes. (3) We validate the effectiveness of our method on multiple mainstream datasets, achieving significant performance improvements.
    \vspace{-15pt}
    \section{Methods}
    \vspace{-10pt} 
    \label{sec:Methods}
    In this section, we will introduce the proposed SCHK-HTC in detail. To enhance the model’s discriminative power for sibling classes by endowing it with domain-specific knowledge, we propose a framework that incorporates both contrastive learning and KG into prompt-tuning. Our architecture's Hierarchical Knowledge-aware Encoder (HK-Encoder) captures intrinsic knowledge hierarchies, while the hierarchical context encoder extracts richly contextualized and highly discriminative features from text. The overall architecture is depicted in Fig.~\ref{fig:1}.
    \vspace{-10pt}
    \subsection{Hierarchical Knowledge-aware Prompt-tuning}
    \vspace{-5pt}
    \label{sec:submethod1}
    \subsubsection{Hierarchical Knowledge-aware Encoder}
    \vspace{-5pt}
    To generate a knowledge-aware representation, we construct a relevant subgraph $\mathcal{G}$ by performing entity linking on the input text against Wikidata \cite{vrandevcic2014wikidata}, extracting the linked entities $\mathcal{E}$ along with their one-hop neighbors and interconnecting relations $\mathcal{R}$.
    The entity linking process is modeled as a two-stage procedure. First, a mention detection(MD) function identifies a set of textual mentions $M=\{m_1, m_2, ..., m_k\}$ within the document $D$. Second, an entity disambiguation step links each mention $m_i$ to its correct entity $e_i^*$ in the KG. This step typically involves generating a set of candidate entities $C(m_i) \subset KG$ and ranking them to find the best match. The final set of linked entities is denoted as $\mathcal{E}=\{e_1,e_2...e_k\}$:
    \vspace{-5pt} 
    \begin{equation}
        \vspace{-5pt}
        \mathcal{E} = \{e_i^* \mid m_i \in \mathrm{MD}(D), e_i^* = \underset{c \in C(m_i)}{\mathrm{argmax}} \, \psi(m_i, c, D)\}
        \vspace{-5pt}
    \end{equation}
    We employ BERT to encode knowledge from two complementary modalities. Given an input sequence $X = \{x_1, x_2, ..., x_n\}$, we concatenate the input text with a pre-defined cloze-style template \lightit{``{\ninept[CLS]} the first layers' knowledge is {\ninept[MASK]}...''} via string concatenation:
    \vspace{-5pt}
    \begin{equation}
        \vspace{-5pt}
        input =template + X
    \end{equation}
    Then we link the entities within $X$ to the subgraph, obtaining a corresponding set of entities $\{e_1, e_2, ..., e_k\}$. For semantic modality, we initialize representations $\{w_1,w_2 ..., w_k\}$ using BERT's embedding layer $Emb_{BERT}$: 
    \vspace{-8pt}  
    \begin{equation}
        \vspace{-5pt} 
         \{w_1,...,w_k\}=Emb_{BERT}(\{e_1,...e_k\})
    \end{equation} 
    For structural modality, we employ a two-stage strategy: initial global embeddings $L$ are generated using Node2Vec \cite{grover2016node2vec} on the subgraph: 
    \vspace{-5pt} 
    \begin{equation}
        L=Node2Vec(\mathcal{E},\mathcal{R})
        \vspace{-5pt}
    \end{equation}
    For each node, we aggregate information from a randomly sampled set of its neighbors in $\mathcal{G}$. This is achieved through random neighbor sampling and feature aggregation, which combines the node's own features with those of its neighbors to produce contextually enriched embeddings. $\mathcal{AGG}$ represents random sample and average aggregation function.
    \vspace{-5pt}
    \begin{equation}
        \{g_1,g_2 ..., g_k\}=\mathcal{AGG}(L,\mathcal{G},\{e_1,e_2...e_k\})
        \vspace{-5pt}
    \end{equation}
    The semantic and structural representations are fused via element-wise addition. Finally, we extract the resulting {\ninept$[MASK]$} token's hidden state from the transformer blocks to serve as the final hierarchical knowledge-aware representation.
\vspace{-12pt}
\subsubsection{Hierarchical Context Encoder}
\label{sec:submethod2}
\vspace{-5pt}
While knowledge-aware features capture entity-specific details, they lack broader sentence context infomation. To complement them, we extract discriminative contextual features using a prompt-based text encoding strategy adapted from DPT \cite{xiong2024dual}.
For each hierarchical layer, we costruct a contrastive prompt \lightit{``{\ninept[CLS]} the first layer is {\ninept[MASK]} rather than \ninept{[MASK]}...''} containing a positive-negative {\ninept$[MASK]$} pair. 
The {\ninept$[MASK]_{pos}$} is assigned the ground-truth label, while the {\ninept$[MASK]_{neg}$} is assigned a confusable sibling label, compelling the model to learn fine-grained distinctions. We define the final-layer feature of the ${\ninept[MASK]_{pos}}$ token as $h_{text}$, which will be utilized in the subsequent fusion stage.
\vspace{-10pt} 
\subsection{Training Objectives}
\vspace{-5pt}
\subsubsection{Knowledge-aware Hierarchical InfoNCE Loss}
\label{sec:submethod3}
\vspace{-5pt}
Our model extracts hierarchical knowledge in a layer-by-layer fashion. To structure the learned representation space, we introduce a Knowledge-aware Hierarchical InfoNCE loss, which is driven by the label hierarchy. The core principle is that for any two samples $x_i$ and $x_j$, let $y_i^{(l)}$ and $y_j^{(l)}$ denote their ground-truth labels at layer $l$. If $y_i^{(l)} = y_j^{(l)}$, then their corresponding knowledge representations, $h_i^{(l)}$ and $h_j^{(l)}$, should exhibit higher similarity than they would with the representation $h_k^{(l)}$ of any sample $x_k$ where the label $y_k^{(l)} \neq y_i^{(l)}$.
This structural constraint is enforced using a contrastive objective. For an anchor sample {\ninept$x_i$} with its layer {\ninept$l$} representation {\ninept$h_i^{(l)}$}, we define the set of positives {\ninept$\mathcal{P}_i^{(l)}$} as samples sharing the label {\ninept$y_i^{(l)}$}, and the set of negatives {\ninept$\mathcal{N}_i^{(l)}$} as those with different labels. The InfoNCE loss for layer $l$ then aims to pull the anchor {\ninept$h_i^{(l)}$} closer to all positive representations {\ninept$\{h^{(l)}_p|p \in \mathcal{P}^{(l)}_i\}$} while pushing it away from all negative representations {\ninept$\{h^{(l)}_n|n \in \mathcal{N}^{(l)}_i\}$}. The loss is formulated as: 
{
    \vspace{-5pt}
    \small
    \begin{equation}
    \label{eq:layer_infonce}
    \mathcal{L}^{(l)}_{\text{K}} = - \log \frac{
        \sum_{p \in \mathcal{P}_i^{(l)}} e^{s(h_i^{(l)}, h_{p}^{(l)}) / \tau}
    }{
        \sum_{p \in \mathcal{P}_i^{(l)}} e^{s(h_i^{(l)}, h_{p}^{(l)}) / \tau} + \sum_{n \in \mathcal{N}_i^{(l)}} e^{s(h_i^{(l)}, h_{n}^{(l)}) / \tau}
    }
    \end{equation}  
}
\vspace{-5pt}
We perform a layer-wise summation of the losses.

{
    \small
    \vspace{-5pt}
    \begin{equation}
        \vspace{-5pt}
        \label{eq:kh-infoNCE}
        \mathcal{L}_{\text{KH-infoNCE}} = \sum_{l=1}^{L} \lambda_l \cdot \mathcal{L}^{(l)}_{\text{K}}
    \end{equation}
}
where {\ninept$s(\cdot)$} represents cosine similarity function, {\ninept$\tau$} is the temperature hyper-parameter, and {\ninept$\lambda_l$} is coefficient per layer.

\vspace{-10pt}
\subsubsection{Sibling Contrastive Learning Loss}
\label{sec:submethod4}
\vspace{-5pt}
To enhance discriminability among sibling classes, we introduce a Sibling Contrastive Learning (SCL) Loss that leverages verbalizers' output for hard-negative mining. For each layer {\ninept$l$}, we select top-k labels with the highest predicted probabilities excluding the ground-truth label from the verbalizer's output as the hard-negative set {\ninept$\mathcal{N}_{hard}^{(l)}$}. These hard negatives are used as targets for a corresponding negative verbalizer in a contrastive objective. The objective of our dual-template contrastive learning strategy is to compel the model to focus on the fine-grained semantic differences between labels, thereby enhancing its discriminative capability. We initialize our verbalizer by first using an LLM to generate detailed textual explanations for each class label. These explanations are subsequently passed through a pre-trained BERT, and we take the resulting ``[CLS]'' token embedding as the initial vectors for our verbalizer. {\ninept$h^{(l)}_{n},h_p^{(l)}$} represent $l$-th layer negative verbalizer and positive verbalizer output respectively, {\ninept$v_{p}^{(l)}$} denotes ground truth label embedding. $v_{n,i}^{(l)}$ is the embedding for the $i$-th hard-negative label sampled at the $l$-th layer. The loss is formulated as:
{
\vspace{-10pt}
\small
\begin{equation}
\label{eq:sibling_infonce}
\begin{aligned}
    \mathcal{L}_{\text{Sibling}} &= -\frac{1}{L} \log\sum_{l=1}^{L} \Bigg( {\frac{s(h_{p}^{(l)},h^{(l)}_{n})}{\tau}} + \\ 
    & \frac{e^{(s(h_{p}^{(l)},v_p^{(l)})/\tau)}}{e^{(s(h_p^{(l)},v_p^{(l)})/\tau)}+\sum_{i=1}^{|\mathcal{N}^{(l)}_{hard}|}e^{(s(h_p^{(l)},v^{(l)}_{n,i}))/\tau}} \Bigg)
\end{aligned} 
\end{equation}
}
\vspace{-20pt}
\subsubsection{Verbalizer Classification Loss}
\vspace{-5pt}
For each hierarchical layer $l$, we fuse the knowledge-aware features {\ninept$h^{(l)}_k$} and textual features {\ninept$h^{(l)}_{\text{text}}$} (from Section \ref{sec:submethod1}) via element-wise addition to form a holistic representation {\ninept$h^{(l)}_{\text{fused}}$}. This fused vector is then projected to logits {\ninept$\mathbf{z}^{(l)}$} over the layer's vocabulary {\ninept$\mathcal{V}^{(l)}$} using a linear verbalizer \cite{schick2020automatically}. Finally, these logits are used to compute the classification loss, employing Binary Cross-Entropy (BCE) for multi-path tasks.
{
    {
        \small
        \vspace{-10pt}
        \begin{equation}
            \mathcal{L}_{\text{BCE}}^{(l)} = - \sum_{j=1}^{|\mathcal{V}^{(l)}|} \left[ y_j^{(l)} \log(\sigma(z_j^{(l)})) + (1 - y_j^{(l)}) \log(1 - \sigma(z_j^{(l)})) \right]
        \end{equation}
    }
}
where $y_j^{(l)}$is the binary ground-truth label for the $j$-th class in layer $l$, and $\sigma(\cdot)$ is the sigmoid function. The negative labels sampled in Section \ref{sec:submethod4} serve as the target labels for the negative verbalizer. For single-path classification, we use the standard Cross-Entropy loss:
\vspace{-5pt}
{
    \small
    \begin{equation}
    \vspace{-5pt}
    \mathcal{L}_{\text{CE}}^{(l)} = - \sum_{j=1}^{|\mathcal{V}^{(l)}|} y_j^{(l)} \log\left(\frac{\exp(z_j^{(l)})}{\sum_{k=1}^{|\mathcal{V}^{(l)}|} \exp(z_k^{(l)})}\right)
    \end{equation}
}
\subsubsection{Objective Function}
\vspace{-5pt}
Overall, final objective is to minimize the weighted combination of classification loss, knowledge-aware infoNCE loss, SCL loss and MLM loss retaining from BERT pre-training. Following HierVerb, we randomly mask 15\% tokens. Final joint loss is formulated as:
{
  \vspace{-5pt}
  \small
  \begin{equation}
    \mathcal{L} = \mathcal{L}_{\text{MLM}} + \mathcal{L}_{\text{BCE/CE}} + \alpha\mathcal{L}_{\text{KH-infoNCE}} + \beta\mathcal{L}_{\text{Sibling}}
  \vspace{-5pt}   
  \end{equation}
}
where $\alpha$ and $\beta$ are hyper-parameters.
\vspace{-10pt}
\section{Experiments and Analysis}
\vspace{-10pt}
\subsection{Experiments Setup}
\vspace{-5pt}
\textbf{Datasets and Evaluation Metrics:}
    We evaluate our method on three standard HTC benchmarks: single-path datasets WOS \cite{kowsari2017hdltex} , DBpedia \cite{sinha2018hierarchical}, and multi-path dataset RCV1-V2 \cite{lewis2004rcv1} . This selection provides diverse hierarchical settings to robustly test our model. Detailed statistics presented in Table~ \ref{tab:datasets} . Similar to previous work,  we measure the experimental results with Macro-F1 and Micro-F1.\\ 
    \noindent
    \textbf{Implementation Details:}
    Both encoders use ``bert-base-uncased'' as their backbone. For the number of randomly sampled neighbors for aggregation, we sample $k=3$. We use Wikidata \cite{vrandevcic2014wikidata} as our KG and set the temperature $\tau=1$ for the KH-InfoNCE loss. The model is trained using the Adam optimizer \cite{kingma2014adam} with a batch size of 8 and a learning rate of $4\times10^{-5}$. The loss balancing parameter $\alpha$ is set to 0.1, $\beta$ is set to 0.2 on WOS and DBpedia, 0.1 on RCV1-V2. We employ an early stopping strategy with a patience of 10 epochs based on the development set's Macro-F1 score. All experiments were conducted on a server with two Intel Xeon Gold 6430 CPUs and one NVIDIA RTX A6000 GPU. 

    \noindent
    \textbf{Baselines:} We compare our method against several strong few-shot HTC baselines: Vanilla-BERT \cite{devlin2019bert}, HGCLR \cite{wang2022incorporating}, HPT \cite{wang2022hpt}, HierVerb \cite{ji2023hierarchical}, and DCL \cite{chen2024retrieval}. Notably, HPT and DCL are considered the current state-of-the-art (SOTA) methods in this domain. These baselines were selected to cover a diverse spectrum of prominent techniques, ranging from the foundational approach of flattening the label hierarchy to more advanced methods like contrastive learning, prompt-tuning, and explicit hierarchical modeling. By benchmarking against these established and varied approaches, including the leading SOTA models, we can rigorously assess the effectiveness of our proposed framework. 
    \begin{table}[htb]
    \vspace{-18pt} 
    \centering
    \small
    \caption{Statistics of the benchmark datasets.}
    \label{tab:datasets}
    \renewcommand{\arraystretch}{0.9}
    \resizebox{\linewidth}{!}{
    \begin{tabular}{ccccccc}
        \toprule
        \textbf{Dataset} & \textbf{Depth} & \textbf{Labels} & \textbf{Avg} & \textbf{Train} & \textbf{Val} & \textbf{Test} \\
        \midrule
        WOS       & 2 & 141 & 2.00 & 30,070  & 7,518  & 9,397    \\
        DBpedia   & 3 & 298 & 3.00 & 277,568 & 30,840 & 34,374   \\
        RCV1-V2   & 4 & 103 & 3.24 & 20,833  & 2,316  & 781,265  \\
        \bottomrule    
    \end{tabular}
    }
    \end{table}
    \vspace{-20pt} 
    {
    \begin{table*}[t]
    \centering
    \renewcommand{\arraystretch}{0.5}
    \caption{F1 scores on 3 datasets under few-shot setting. \textbf{Bold}: best results. The dagger (\dag) indicates the direct  utilization of results from \cite{ji2023hierarchical} . Our implementation results are marked by ``*'' . We modify negative sampling strategy of DCL to strictly adhere to the k-shot setting. We report the mean F1 scores (\%) over 5 random seeds.
        }
    \label{tab:table2}
    \resizebox{\linewidth}{!}{
    \begin{tabular}{c l cc cc cc} 
        \toprule
        & & \multicolumn{2}{c}{\textbf{WOS}} & \multicolumn{2}{c}{\textbf{DBpedia}} & \multicolumn{2}{c}{\textbf{RCV1-V2}} \\
        \cmidrule(lr){3-4} \cmidrule(lr){5-6} \cmidrule(lr){7-8}
        \textbf{Shot} & \textbf{Method} & Micro-F1 & Macro-F1 & Micro-F1 & Macro-F1 & Micro-F1 & Macro-F1 \\
        \midrule
        
        \multirow{8}*{1} 
        & Vanilla-BERT \dag  & $2.99 \pm 20.85\,(5.12)$  & $0.16 \pm 0.10\,(0.24)$   & $14.43 \pm 13.34\,(24.27)$ & $0.29 \pm 0.01\,(0.32)$ & $7.32 \pm 10.33\,(9.32)$ & $3.73 \pm 0.10\,(3.73)$ \\
        & HGCLR \dag         & $9.77 \pm 11.77\,(16.32)$ & $0.59 \pm 0.10\,(0.63)$   & $15.73 \pm 31.07\,(25.13)$ & $0.28 \pm 0.10\,(0.31)$ & $26.46  \pm 1.27\,(26.80)$ & $1.34 \pm 0.93 \,(1.71)$ \\
        & HPT \dag           & $50.05 \pm 6.80\,(50.96)$ & $25.69 \pm 3.31\,(27.76)$ & $72.52 \pm 10.20\,(73.47)$ & $31.01 \pm 2.61\,(32.50)$ & $27.70 \pm 5.32 \,(28.51)$ & $3.35 \pm 2.22 \,(3.90)$ \\
        & HierVerb *         & $58.87 \pm 4.38\,(61.13)$ & $44.28 \pm 4.86\,(47.84)$ & $91.66 \pm 0.08\,(91.71)$  & $85.32 \pm 0.04\,(85.44)$ & $40.85 \pm 3.12\,(41.22)$ & $4.82 \pm 1.71\,(5.71)$ \\
        & DCL *              & $58.34 \pm 6.41\,(60.84)$ & $44.14 \pm 5.68\,(47.66)$ & $91.68 \pm 0.04\,(91.77)$  & $85.28 \pm 0.08\,(85.44)$ & $40.75 \pm 3.28\,(41.01)$ & $4.76 \pm 2.01\,(5.66)$ \\
        & SCHK-HTC(Ours)     & $\textbf{59.65} \pm \textbf{5.62}\,\textbf{(61.78)}$  & $\textbf{45.04} \pm \textbf{5.12}\,\textbf{(48.15)}$ & $\textbf{91.91} \pm \textbf{0.11}\,\textbf{(91.97)} $ & $\textbf{85.41} \pm \textbf{0.08}\,\textbf{(85.48)}$ & $\textbf{40.97} \pm \textbf{4.13}\,\textbf{(41.34)}$ & $\textbf{4.93} \pm \textbf{2.01}\,\textbf{(5.92)}$ \\ 
        \midrule
            
        \multirow{8}{*}{2}
        & Vanilla-BERT \dag & $46.31 \pm 0.65\,(46.85)$ & $5.11 \pm 1.31\,(5.51)$  & $87.02 \pm 3.89\,(88.20)$  & $69.05 \pm 26.81\,(73.28)$ & $8.07 \pm 2.18\,(9.13)$  & $2.76 \pm 6.01\,(4.11)$   \\
        & HGCLR \dag        & $45.11 \pm 5.02\,(47.56)$ & $5.80 \pm 11.63\,(9.63)$ & $90.32 \pm 0.64\,(91.11)$  & $71.46 \pm 0.17\,(71.78)$ & $34.33 \pm 4.81\,(37.28)$ & $2.51 \pm 6.12\,(6.12)$ \\
        & HPT \dag          & $57.45 \pm 1.89\,(58.99)$ & $35.97 \pm 11.89\,(39.94)$ & $93.71 \pm 0.01\,(93.87)$ & $81.12 \pm 1.33\,(82.42)$ & $38.93 \pm 3.55\,(40.47)$ & $8.31 \pm 5.26\,(10.52)$ \\
        & HierVerb *        & $64.43 \pm 5.39\,(66.89)$ & $54.04 \pm 3.24\,(56.69)$ & $93.21 \pm 0.01\,(93.22)$  & $88.96 \pm 0.02\,(89.02)$ & $48.00 \pm 2.27\,(49.21)$ & $11.74 \pm 1.85 \,(12.69)$ \\
        & DCL *             & $63.26 \pm 4.61\,(66.27)$ & $53.56 \pm 3.87\,(56.16)$ & $92.61 \pm 0.16\,(92.85)$  & $88.27 \pm 0.09\,(88.49)$ & $47.56 \pm 1.86\,(48.99)$ & $11.54 \pm 1.87\,(12.44)$ \\
        & SCHK-HTC(Ours)    & $\textbf{66.12} \pm \textbf{4.34}\,\textbf{(67.36)}$ & $\textbf{54.38} \pm \textbf{2.89}\,\textbf{(56.88)}$ & $\textbf{93.24} \pm \textbf{0.02} \,\textbf{(93.25)}$  & $\textbf{89.07} \pm \textbf{0.04} \,\textbf{(89.08)}$ & $\textbf{48.06} \pm \textbf{2.09} \,\textbf{(49.35)} $ & $ \textbf{11.88} \pm \textbf{2.43} \,\textbf{(12.88)}$  \\
        \midrule
        
        \multirow{8}{*}{4}  
        & Vanilla-BERT \dag     & $56.00 \pm 4.25\,(57.18)$ & $31.04 \pm 16.65\,(33.77)$ & $92.94 \pm 0.66\,(93.38)$ & $84.63 \pm 0.17\,(85.47)$ &$17.94 \pm 0.01\,(18.00)$ &$1.45 \pm 0.01\,(1.57)$\\
        & HGCLR \dag            & $56.80 \pm 4.24\,(57.96)$ & $32.34 \pm 15.39\,(33.76)$ & $93.14 \pm 0.01\,(93.22)$ & $84.74 \pm 0.11\,(85.11)$ &$45.53 \pm 4.20\,(47.71)$ &$8.56 \pm 1.63\,(9.92)$\\
        & HPT \dag              & $65.57 \pm 1.69\,(67.06)$ & $45.89 \pm 9.78\,(49.42)$  & $94.34 \pm 0.28\,(94.83)$ & $90.09 \pm 0.87\,(91.12)$ &$52.62 \pm 0.20\,(52.73)$ &$20.01 \pm 0.31\,(20.21)$\\
        & HierVerb *            & $72.37 \pm 0.83\,(73.64)$ & $62.49 \pm 1.48\,(64.47)$  & $94.51 \pm 0.13\,(94.88)$ & $90.77 \pm 0.33\,(91.43)$ &$ \textbf{56.86} \pm \textbf{0.44}\,\textbf{(57.11)}$ &$\textbf{22.07} \pm \textbf{0.32}\,\textbf{(22.42)}$\\
        & DCL *                 & $72.44 \pm 0.99\,(73.52)$ & $63.06 \pm 1.07\,(64.88)$  & $94.32 \pm 0.20\,(94.83)$  & $90.22 \pm 0.22\,(90.99) $ & $56.44 \pm 0.39\,(56.97)$ & $21.77 \pm 0.27\,(22.14)$ \\
        & SCHK-HTC(Ours)        & $\textbf{72.88}\pm \textbf{1.09} \,\textbf{(73.81)}$ & $\textbf{63.81} \pm \textbf{1.24} \,\textbf{(65.24)}$ & $\textbf{94.83} \pm \textbf{0.15} \,\textbf{(95.16)}$  & $\textbf{90.89} \pm \textbf{0.12}\,\textbf{(91.57)}$ & $ 56.66 \pm 0.34 \,(57.03) $ & $ 21.89 \pm 0.26 \,(22.27)$       \\
        \midrule
        \multirow{8}{*}{8}  & Vanilla-BERT \dag      & $66.24 \pm 1.96\,(67.53)$ & $50.21 \pm 5.05\,(52.60)$ & $94.39 \pm 0.06\,(94.57)$ & $87.63 \pm 0.28\,(87.78)$ &$57.27 \pm 0.04\,(57.51)$ &$23.93 \pm 0.45\,(24.46)$\\
                            & HGCLR \dag             & $68.34 \pm 0.96\,(69.22)$ & $54.41 \pm 2.97\,(55.99)$ & $94.70 \pm 0.05\,(94.94)$ & $88.04 \pm 0.25\,(88.61)$  &$58.90 \pm 1.61\,(60.30)$ &$27.03 \pm 0.20\,(27.41)$\\
                            & HPT \dag               & $78.22 \pm 0.99\,(77.23)$ & $67.20 \pm 1.89\,(68.63)$ & $95.49 \pm 0.01\,(95.57)$ & $92.35 \pm 0.03\,(92.52)$  &$59.92 \pm 4.25\,(61.47)$ &$29.03 \pm 6.23\,(32.19)$\\
                            & HierVerb *             & $78.12 \pm 0.55\,(78.87)$ & $69.60 \pm 0.91\,(70.56)$ & $95.19 \pm 0.04\,(95.22)$ & $92.44 \pm 0.01\,(92.51)$  &$63.76 \pm 2.22\,(64.71)$ &$\textbf{31.13} \pm \textbf{1.63}\,\textbf{(32.52)}$\\
                            & DCL *                  & $77.97 \pm 0.46\,(78.21)$ & $69.78 \pm 0.77\,(70.88)$ & $95.52 \pm 0.03\,(95.56)$ & $92.43 \pm 0.02\,(92.52)$ & $63.81 \pm 1.26\,(64.44)$ & $30.24 \pm 1.58\,(31.99)$   \\
                            & SCHK-HTC(Ours)         & $\textbf{78.34} \pm \textbf{0.33}\,\textbf{(78.89)}$ & $\textbf{70.06} \pm \textbf{0.34}\textbf{(71.13)}$ & $\textbf{95.78} \pm \textbf{0.04} \,\textbf{(95.79)}$ & $\textbf{92.46} \pm \textbf{0.01}\,\textbf{(92.55)} $ & $\textbf{63.97} \pm \textbf{1.83}\,\textbf{(65.02)}$ & $30.88 \pm 1.66\,(32.17)$  \\
        \midrule
        \multirow{8}{*}{16} & Vanilla-BERT \dag & $75.52 \pm 0.32\,(76.07)$ & $65.85 \pm 1.28\,(66.96)$ & $95.31 \pm 0.01\,(95.37)$ & $89.16 \pm 0.07\,(89.35)$ & $63.68 \pm 0.01\,(64.10)$ & $34.00 \pm 0.67\,(34.41)$ \\
                            & HGCLR \dag        & $76.93 \pm 0.52\,(77.46)$ & $67.92 \pm 1.21\,(68.66)$ & $95.49 \pm 0.04\,(95.63)$ & $89.41 \pm 0.09\,(89.71)$ & $63.91 \pm 1.42\,(64.81)$ & $33.25 \pm 0.10\,(33.50)$ \\
                            & HPT \dag          & $79.85 \pm 0.41\,(80.58)$ & $72.02 \pm 1.40\,(73.31)$ & $96.13 \pm 0.01\,(96.21)$ & $93.28 \pm 0.06\,(93.49)$ & $65.45 \pm 0.80\,(66.03)$ & $36.12 \pm 0.20\,(36.23)$ \\
                            & HierVerb *        & $80.35 \pm 0.22\,(81.17)$ & $73.56 \pm 0.12\,(73.68)$ & $96.17 \pm 0.01\,(96.21)$ & $93.28 \pm 0.06\,(93.49)$ & $\textbf{65.50} \pm \textbf{1.41}\,\textbf{(66.59)}$ & $35.84 \pm 1.21\,(36.12)$\\
                            & DCL *             & $80.66 \pm 0.26\,(81.09)$ & $73.66 \pm 0.17\,(73.79)$ & $96.20 \pm 0.02\,(96.24$) & $93.26 \pm 0.02 \,(93.52)$ & $65.46 \pm 0.95\,(66.21)$ & $\textbf{36.16} \pm \textbf{0.79}\,\textbf{(36.45)}$ \\
                            & SCHK-HTC(Ours)    & $\textbf{81.31} \pm \textbf{0.33} \,\textbf{(81.36)}$ & $\textbf{74.02} \pm \textbf{0.22}\,\textbf{(74.15)}$ & $\textbf{96.26} \pm \textbf{0.02} \,\textbf{(96.28)}$ & $\textbf{93.29} \pm \textbf{0.07}\,\textbf{(93.55)} $ & $65.48 \pm 1.13\, (66.44) $ & $36.08 \pm 0.88 \,(36.31)$ \\
        \bottomrule
    \end{tabular}
    }
    \vspace{-10pt}
    \end{table*}
    
    }
    \subsection{Main Results}
    \vspace{-5pt} 
     Table ~\ref{tab:table2} summarizes the results of our comprehensive evaluation. To ensure a fair and direct comparison, we re-implemented key baselines within a unified experimental setup. The results clearly indicate that our method consistently outperforms competing approaches across the majority of k-shot settings, demonstrating particularly significant gains on the WOS and DBpedia datasets. This suggests that our approach excels in addressing single-path tasks. While its performance advantage on RCV1-V2 diminishes as the number of shots increases, it still maintains a competitive edge in extremely low-shot scenarios. This performance disparity can be attributed to the geometric properties targeted by our contrastive learning objective. The model's superior performance in single-path tasks demonstrates its effectiveness in shaping distinct, well-separated class clusters, which is the primary strength of contrastive learning. However, this very mechanism becomes less optimal in multi-path scenarios. Here, co-occurring ground-truth labels can exert conflicting optimization ``pulls'' on a sample's representation, making it more challenging to form sharp decision boundaries. Although the HK-Encoder introduces an additional view to incorporate knowledge graph information, our experimental results indicate that this does not adversely affect the training convergence speed.
    \vspace{-10pt} 
    
    \subsection{Embedding Visualization}
\begin{figure}[htbp]
    \vspace{-10pt}
    \centering
    \begin{subfigure}[b]{0.15\textwidth}
        \includegraphics[width=\linewidth]{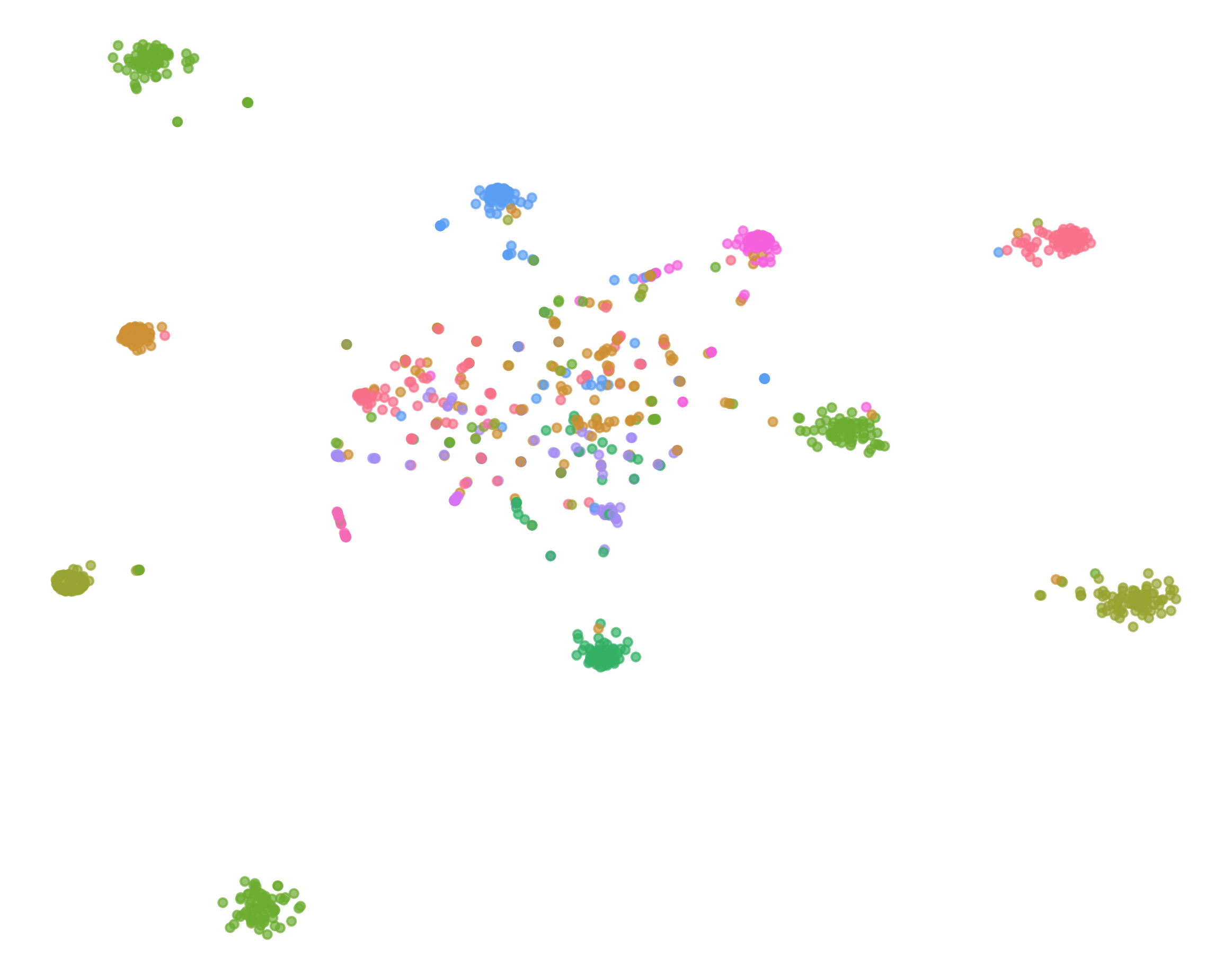}
        \caption{HierVerb}
        \label{fig:1a}
    \end{subfigure}
    \hspace{30pt} 
    \begin{subfigure}[b]{0.15\textwidth}
        \includegraphics[width=\linewidth]{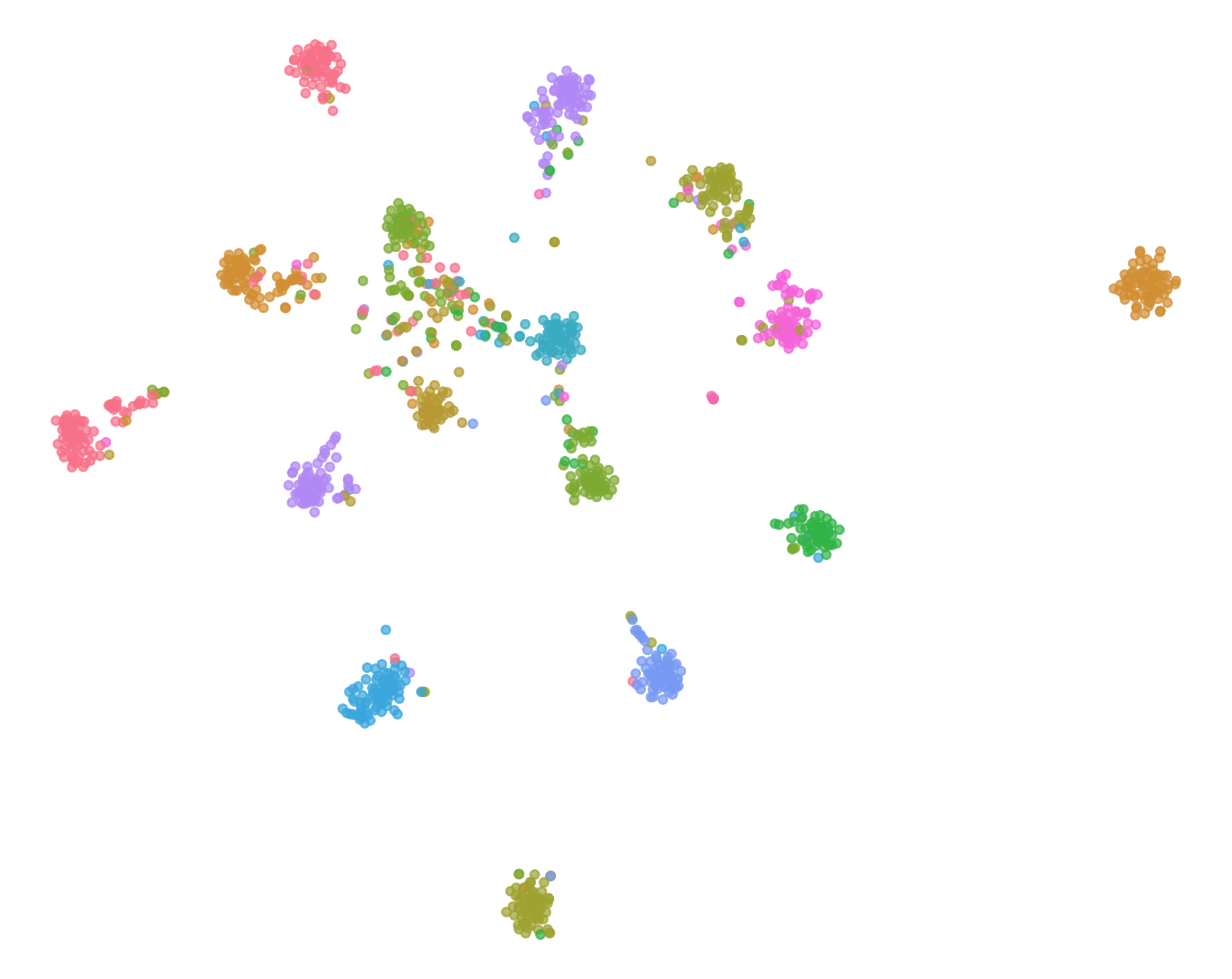}
        \caption{Ours}
        \label{fig:1b}
    \end{subfigure}
    \vspace{-10pt}
    \caption{T-SNE visualization of label representations on WOS.}
    \vspace{-15pt}
    \label{fig:combined}
    \end{figure}

    \vspace{-5pt}  
    Figure~\ref{fig:combined} compares the t-SNE projections of label embeddings from our method and HierVerb on the WOS 16-shot task. The learned embedding distributions for sub-categories of the CS class are depicted in these two subfigures. Our method's superiority is most evident in its ability to handle difficult samples, where it effectively resolves the semantic ambiguity between sibling classes. While HierVerb's feature clusters for these samples appear diffuse and overlapping, our approach maintains clear and well-defined cluster boundaries, demonstrating its enhanced discriminability. 
    \vspace{-8pt}
    \subsection{Ablation Study}
    \vspace{-5pt}
    To rigorously assess the individual contribution of each component within our SCHK-HTC framework, we conducted a comprehensive ablation study. We systematically dismantled the full model by individually removing three key modules: HK-Encoder, HK-InfoNCE loss, and SCL loss. The performance of these ablated variants was evaluated against the full model on the WOS dataset. As detailed in Table \ref{table:ablation_wos}, we report both standard metrics Micro-F1, Macro-F1 and additional path-constrained metrics C-MicroF1, C-Macro-F1 from \cite{yu2022constrained} to provide a multi-faceted view of the impact.
    Removing the HK-Encoder slightly degrades performance, confirming the benefit of our knowledge-aware feature extraction. The decline is more substantial in path-constrained metrics when the HK-InfoNCE loss is removed, highlighting its key role in injecting hierarchical structure. Most notably, performance degrades most sharply without the SCL loss. This validates that our dual-template contrastive learning is effective at resolving semantic ambiguity among sibling classes.
    \vspace{-10pt}
    \begin{table}[h]
        \centering
        \small 
        \renewcommand{\arraystretch}{0.6} 
        \vspace{-15pt}
        \caption{Ablation study on WOS dataset}
        \label{table:ablation_wos} 
        \resizebox{\linewidth}{!}{%
        \begin{tabular}{clllll}
        \toprule
        \textbf{K} & \textbf{Ablation Models} & \multicolumn{4}{c}{\textbf{WOS}}  \\
        \cmidrule(lr){3-6}
        & & Micro-F1 & Macro-F1 & C-MicroF1 & C-MacroF1 \\
        \midrule

        \multirow{5}{*}{1} 
        & Ours                                               & \textbf{59.65}                     & \textbf{45.04}                     & \textbf{56.18}               & \textbf{41.05} \\
        & \textit{r.m.} KH-infoNCE loss                      & 58.13 \scriptsize{-2.55\%}         & 44.63 \scriptsize{-0.91\%}         & 51.86 \scriptsize{-7.79\%}   & 31.84 \scriptsize{-22.44\%}\\
        & \textit{r.m.} KH-Encoder                           & 57.88 \scriptsize{-2.97\%}         & 44.27 \scriptsize{-1.71\%}         & 52.04 \scriptsize{-7.37\%}   & 32.13 \scriptsize{-21.73\%}\\
        & \textit{r.m.} SCL loss                             & 58.35 \scriptsize{-2.18\%}         & 44.48 \scriptsize{-1.24\%}         & 52.92 \scriptsize{-5.80\%}   & 37.46 \scriptsize{-8.74\%}\\
        \midrule

        \multirow{5}{*}{2} 
        & Ours                                               & \textbf{66.12}                     & \textbf{54.38}                     & \textbf{62.99}               &  \textbf{50.13} \\
        & \textit{r.m.} KH-infoNCE loss                      & 65.40 \scriptsize{-1.09\%}         & 53.89 \scriptsize{-0.90\%}         & 57.61 \scriptsize{-8.54\%}   & 46.27 \scriptsize{-6.03\%}\\
        & \textit{r.m.} KH-Encoder                           & 65.87 \scriptsize{-0.38\%}         & 53.94 \scriptsize{-0.81\%}         & 57.88 \scriptsize{-8.11\%}   & 46.35 \scriptsize{-5.87\%}\\
        & \textit{r.m.} SCL Learning                         & 65.23 \scriptsize{-1.35\%}         & 53.47 \scriptsize{-1.67\%}         & 60.32 \scriptsize{-4.34\%}   & 48.59 \scriptsize{-1.32\%}\\
        \midrule

        \multirow{5}{*}{4} 
        & Ours                                               & \textbf{72.88}                     & \textbf{63.81}                     & \textbf{72.23}               & \textbf{61.37} \\
        & \textit{r.m.} KH-infoNCE loss                      & 72.51 \scriptsize{-0.51\%}         & 62.70 \scriptsize{-1.74\%}         & 69.24 \scriptsize{-4.14\%}   & 56.09 \scriptsize{-8.60\%}\\
        & \textit{r.m.} KH-Encoder                           & 72.05 \scriptsize{-1.14\%}         & 62.52 \scriptsize{-2.02\%}         & 69.95 \scriptsize{-3.17\%}   & 55.84 \scriptsize{-9.01\%}\\
        & \textit{r.m.} SCL Learning                         & 72.22 \scriptsize{-0.91\%}         & 62.22 \scriptsize{-2.49\%}         & 71.41 \scriptsize{-1.15\%}   & 59.67 \scriptsize{-2.77\%}\\
        \bottomrule
        \end{tabular}%
        }
        \vspace{-10pt}
        \end{table}
    \vspace{-15pt}
    \subsection{Deeper Layer Classification Analysis}
    \vspace{-5pt}
    To specifically quantify our model's ability to resolve ambiguity among deep-level sibling labels, we perform a targeted evaluation. We report the classification accuracy (Acc\%) on the deepest level of WOS and DBpedia dataset in Figure \ref{fig:Deeper Level}. This metric provides a direct measure of our model's discriminative power where it matters most, highlighting the efficacy of our methods.
    \vspace{-10pt}
    \section{Conclusion}
    \vspace{-5pt}
    This paper proposes the SCHK-HTC framework to address the challenge of distinguishing between similar sibling labels in few-shot HTC. Our core contribution is a novel mechanism that integrates hierarchical knowledge via a prompt-based encoder and dual-template prompt-tuning to facilitate SCL. This approach alleviates the suboptimal classification performance common in few-shot scenarios by fostering more discriminative feature representations. Experiments validate its efficacy, showing that our model not only significantly outperforms SOTA methods overall but also demonstrates unparalleled success in accurately classifying the most fine-grained and challenging labels at the lowest levels of the hierarchy, directly addressing sibling label confusion.
    \label{sec:refs}    
    { 
        \clearpage 
	\small
	\begin{spacing}{1.0}
            \bibliographystyle{IEEEtran}
		\bibliography{inproceedings,article}
	\end{spacing} 
    } 
\end{document}